\documentclass[runningheads]{llncs}
\usepackage{graphicx}

\usepackage{amsmath,amssymb} % define this before the line numbering.
\usepackage{color}

\usepackage{times}
\usepackage{epsfig}
\usepackage{booktabs} % For formal tables

\usepackage{import}
\usepackage{epstopdf}
\usepackage{caption}
\usepackage{subcaption}
\usepackage{tabu}
\usepackage{tabularx}
\usepackage{multirow}
\usepackage{xfrac}
\usepackage{comment}

\usepackage{makecell}
\usepackage{xspace}
\usepackage{hyperref}
\hypersetup{
	colorlinks=true,
	linkcolor=blue,
	filecolor=magenta,
	urlcolor=cyan,
}

\usepackage[square,numbers]{natbib}

\newcommand*{\eg}{e.g.\@\xspace}
\begin{document}
%===========================================================

\title{Spatio-Temporal Fusion Networks for Action Recognition\thanks{This work was supported in part by the National Science Foundation under grant IIS-1212948.}} % Replace your paper's title here
\titlerunning{Spatio-Temporal Fusion Networks} % Replace an abstracted version of your paper's title here

%===========================================================

\author{Sangwoo Cho \and Hassan Foroosh} %\and  \inst{1}
%Third Author\inst{3}\orcidID{2222--3333-4444-5555}}
%
%Please include author names in full in the paper, 
%If any authors have names that can be parsed into FirstName LastName in multiple ways, please include the correct parsing, in a comment to the volume editors:
%\index{Lastnames, Firstnames}

\authorrunning{Cho et al.} % A shorter version of authors' name

%===========================================================

\institute{University of Central Florida, Orlando FL 32816, USA \\
\email{swcho@knights.ucf.edu, foroosh@cs.ucf.edu}}

\maketitle

%===========================================================
\begin{abstract}
The video based CNN works have focused on effective ways to fuse appearance and motion networks, but they typically lack utilizing temporal information over video frames. In this work, we present a novel spatio-temporal fusion network (STFN) that integrates temporal dynamics of appearance and motion information from entire videos. The captured temporal dynamic information is then aggregated for a better video level representation and learned via end-to-end training.
The spatio-temporal fusion network consists of two set of Residual Inception blocks that extract temporal dynamics and a fusion connection for appearance and motion features.
The benefits of STFN are: (a) it captures local and global temporal dynamics of complementary data to learn video-wide information; and (b) it is applicable to any network for video classification to boost performance.
We explore a variety of design choices for STFN and verify how the network performance is varied with the ablation studies.
We perform experiments on two challenging human activity datasets, UCF101 and HMDB51, and achieve the state-of-the-art results with the best network.

\keywords{Action Recognition \and Spatio-Temporal Fusion \and Temporal Dynamics.}
\end{abstract}

%%%%%%%%% BODY TEXT
\section{Introduction} \label{introduction}

\begin{figure}[!htb]
	\begin{center}
		\includegraphics[width=0.7\linewidth]{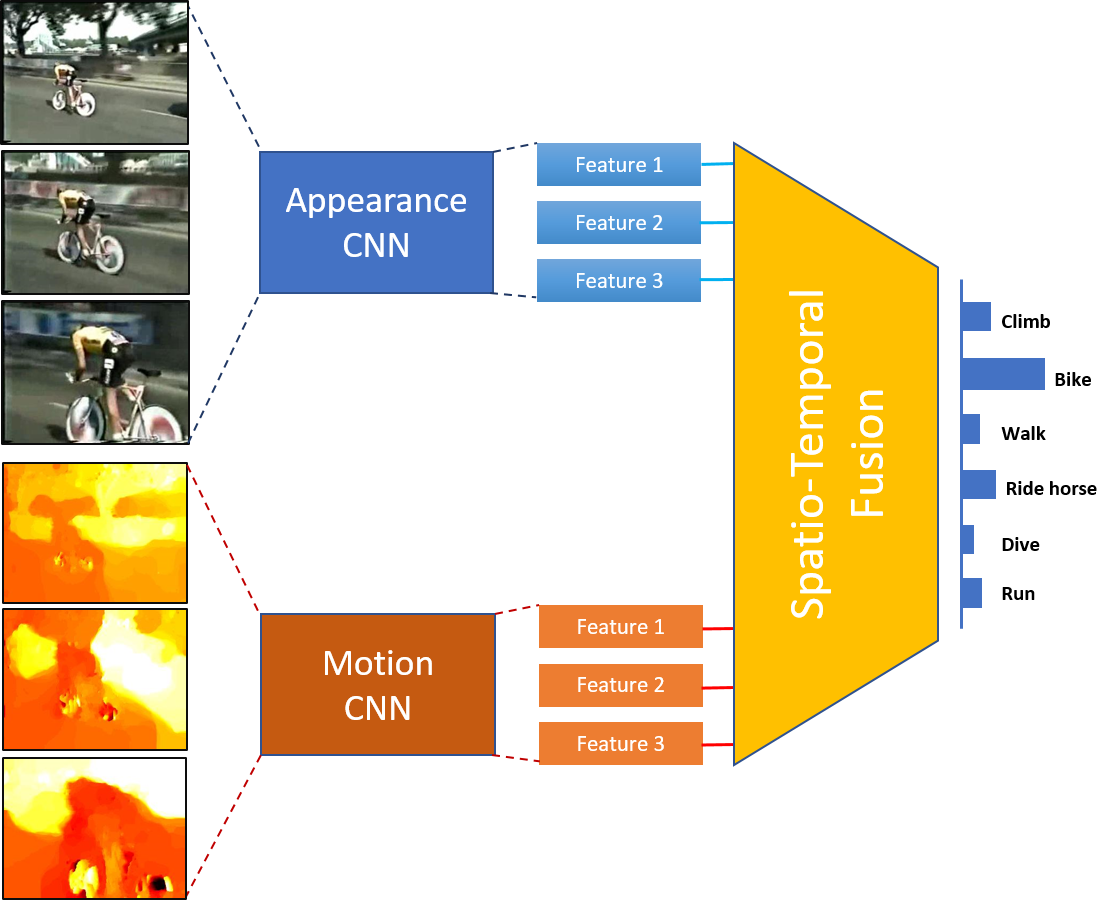}
	\end{center}
	\caption{
		An illustration of spatio-temporal fusion network for action recognition. Given multiple segments of a video, the network extracts temporal dynamics of appearance and motion cues and fuses them to build a spatio-temporal video representation via end-to-end learning. The appearance and motion ConvNets share the same weights and are employed to extract appearance and motion features, respectively.		
	}
	\label{fig:detail_desc}
\end{figure}

Video-based action recognition is an active research topic due to its important practical applications in many areas, such as video surveillance, behavior analysis, and human-computer interaction. Unlike a single image that contains only spatial information, a video provides additional motion information as an important cue for recognition. Although a video provides more information, it is non-trivial to extract the information due to a number of difficulties such as viewpoint changes, camera motions, and scale variations, to name a few. It is thus crucial to design an effective and generalized representation of a video.

Convolutaional Neural Networks (ConvNets) \cite{intro_ref1} have been playing a key role in solving hard problems in various areas of computer vision, \eg image classification \cite{intro_ref1,related_ref1,intro_ref2} and human face recognition \cite{intro_ref4}. ConvNets also have been employed to solve the problem of action recognition \cite{ar_ref1,ar_ref12,ar_ref13,ar_ref14,survey_paper} in recent literature. 
The data-driven supervised learning enables to achieve discriminating power and proper representation of a video from raw data. However, ConvNets for action recognition have not shown a significant performance gain over the methods utilizing hand-crafted features \cite{feat_ref3,related_ref7}. We speculate that the main reason for the lack of big impact is that ConvNets employed in action recognition do not take full advantage of temporal dynamics among frames. 

A two-stream \cite{ar_ref1} ConvNet is one of the popular approaches used in action recognition utilizing the appearance and motion data. However, the two data streams are typically trained with separate ConvNets and only combined by averaging the prediction scores. This approach is not helpful when the two information are needed simultaneously, \eg motions of brushing teeth and brushing hair are similar so appearance information is needed to discriminate them. Due to the lack of spatio-temporal feature for action recognition, several methods \cite{ar_ref19,st-arts4,st-arts7} attempted to incorporate both sources of information. They typically take frame-level features and integrate them using an RNN \cite{related_ref1} network and temporal feature pooling \cite{ar_ref5,st-arts2,st-arts1} in order to incorporate temporal information.
However, they still lack of extracting a representation that captures video-wide temporal information.

In this work, we aim to investigate a proper model to fuse the appearance and motion dynamics to learn video level spatio-temopral representation. 
STFN aggregates different size of local temporal dynamics in multiple video segments and combines them to obtain video level spatio-temporal representation.
STFN is mainly motivated by two components: a residual-inception module \cite{ar_ref14}, and 1D convolution layers \cite{ta2}. The former is suitable for extracting latent features and the latter works well in extracting temporal dynamics. We modify the original residual-inception module \cite{ar_ref14} and design a new block for spatio-temporal fusion.
The new residual-inception block processes local and global temporal dynamics for each data. Given the extracted dynamic information, appearance and motion dynamics are merged with fusion operations for spatio-temporal features.
%We first encode the spatial and temporal features as a high dimensional vector space and represent a video by concatenating the vector representations. 
%This video representation is well-suited with our proposed Residual Inception blocks extracting high level feature representations. 
%We present a novel Temporal Fusion Network (TFN) that extracts high level feature representations and combines spatial and temporal information by connecting two signals. 
This method overcomes the previous drawback, i.e. the lack of utilizing video-wide temporal information, and learning spatio-temporal features.
We investigate a variety of different fusion methods and execute ablation studies to find the best networks.
%Empirically, we find STFN is robust to fusion an asymmetric and bidirection fusion methods are both useful for different datasets. 

Our key contributions can thus be summarized as follows: 
(i) A convolution block, effective to extract temporal representations, is proposed. (ii) A novel ConvNet is introduced to learn spatio-temporal features effectively by fusing two different features properly. (iii) STFN achieves state-of-the-art performance on the two challenging datasets, UCF101 (95.4$\%$) and HMDB51 (72.1$\%$). (iv) The entire system is easy to implement and is trained by an end-to-end learning of deep networks.

The rest of this paper is organized as follows. In Section \ref{relatedwork}, we discuss related works. We describe our proposed method in Section \ref{approach}. Experimental results and analysis are presented in Section \ref{experiments}. Finally, we conclude our work in Section \ref{conclusion}.
%------------------------------------------------------------------------
\section{Related Work} \label{relatedwork}

Several works using ConvNets to acquire temporal information for action recognition have been studied. In  \cite{ar_ref7}, hand crafted features are used in the pooling layer of ConvNet to take advantage of both merits of hand-designed and deep learned features. Temporal information from optical flow is explicitly learned with ConvNets in  \cite{ar_ref1} and the result is fused with the effect of the trained spatial (appearance) ConvNet. \cite{ar_ref5} connects several convolution layers of two stream ConvNets to capture spatio-temporal information. Although the aforementioned approaches capture temporal information in small time windows, they fail to capture long-range temporal sequencing information that contains long-range ordered information. 

Several works modeling a video-level representation or modeling long temporal information with ConvNets have also been investigated. \cite{ar_ref6} proposes a method that employs a ranking function to generate a video-wide representation that captures global temporal information. In  \cite{related_ref8}, a HMM model is used to capture the appearance transitions and a max-margin method is employed for temporal information modeling in a video. \cite{ar_ref8,related_ref8,related_ref9} utilize LSTM \cite{related_ref1} unit in their ConvNets and attempt to capture long-range temporal information. However, the most natural way of representing a video as long-range ordered temporal information is not fully exploited.

Recently several researches \cite{ta1,ta2} have used frame level representations for predicting actions with temporal ConvNets. The rational behind these methods is to extract the temporal dynamics more directly by utilizing 1D convolution over time. This approach is widely used in a sentence classification \cite{text_ref1,text_ref3,text_ref4} problem in Natural Language Processing literature. Each word is encoded to vectors and 1D convolution over a sequence of words extracts semantic information between words. For videos, two stream \cite{ar_ref1} ConvNets are typically employed to train appearance and motion features separately. Once the two streams are trained, sampled RGB or optical flow video frames are fed to each network to extract appearance and motion features respectively. This is the standard feature extraction method and each frame can be represented in a vector form. The biggest advantage of the feature representation is that the temporal information distributed over entire videos can be effectively extracted by using 1D convolutions. Our work is based on the 1D convolution layers to obtain temporal dynamics of appearance and motion cues.

Many ConvNets \cite{ar_ref2,ar_ref9,ar_ref11,ar_ref13} for image recognition are utilized for action recognition as well. Among them, a concept of the inception is useful to our encoded data to extract more informative features. The encoded features are convoluted over time with different kernel sizes and concatenated. This process extracts local and global temporal information similar to extracting N-gram semantic information in NLP. \cite{ar_ref14} introduces an effective residual inception module, which basically has another shortcut connection to the inception module. We employ the residual inception module with 1D convolution layers as it is suitable for extracting temporal dynamics.

The critical drawback of the two-stream \cite{ar_ref1} ConvNets is the two features cannot be integrated in feature level. In order to solve this problem, different fusion methods are introduced. In \cite{ar_ref19} they try to extract spatio-temporal features directly by applying 3D convolution to a stack of input frames. \cite{st-arts4,st-arts5} connect learned two stream ConvNets to integrate the two stream signals generating the spatio-temporal features. \cite{st-arts7} encodes local deep features as a super vector efficiently so that spatio-temporal information can be handled with spatio-temporal ConvNets. We utilize different basic fusion operations, average, maximum, and multiply, as investigated in \cite{st-arts1,st-arts4,st-arts5}. Since we combine the appearance and motion features, we naturally take advantage of two stream ConvNet architecture and connect them with different fusion methods. This work provides a systematic investigation of fusion methods and ablation studies to choose the best fusion methods for better performance.
%------------------------------------------------------------------------
\section{Approach} \label{approach}

\begin{figure}[tb]
	\begin{center}
		\includegraphics[width=0.8\linewidth]{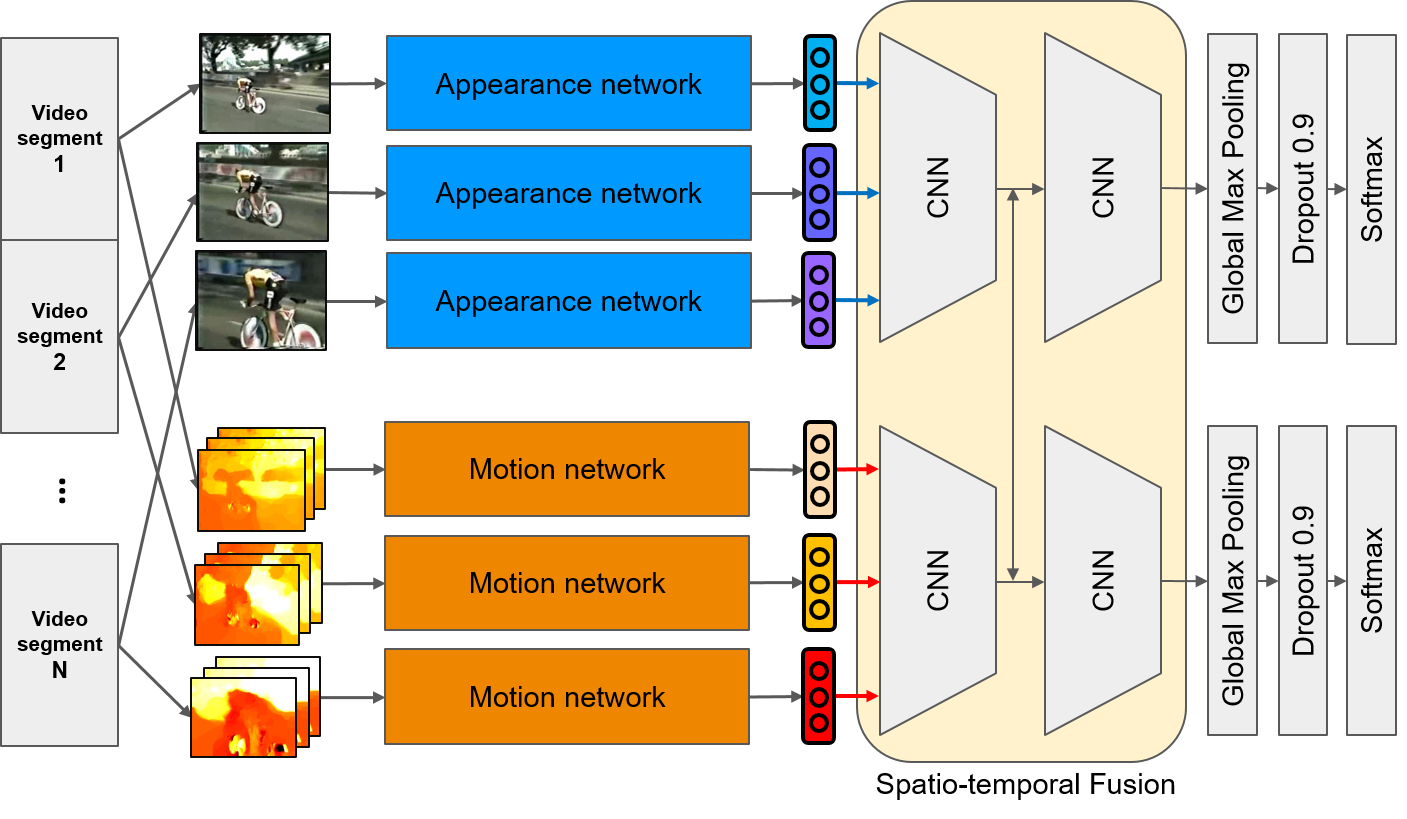}
	\end{center}
	\caption{
		The proposed spatio-temporal fusion network. The number of segments is an arbitrary number. We use three segments in the figure for illustration purpose.
	}
	\label{fig:STFN}
\end{figure}

A video contains many redundant temporal information between consecutive frames. Instead of densely sampled feature points \cite{ar_ref16}, \cite{SOTA3} samples frames in different video segments, while \cite{ar_ref1} deals with multiple consecutive frames. These techniques train ConvNets for different modalities, appearance and motion, and use late fusion to combine them.
However, two issues are raised from these methods: (1) multiple consecutive frames only cover local temporal dynamics not global temporal dynamics over videos, and (2) the prediction score fusion only captures dynamic of each appearance and motion cue separately not the spatio-temporal dynamics.
In this section, we propose a spatio-temporal fusion network (STFN) to extract temporal dynamic information over an entire video and combine appearance and motion dynamics, using end-to-end ConvNets training, as shown in Fig. \ref{fig:STFN}.
The network has the following properties: (1) convolutions are computed over time so that the temporal dynamic information is extracted; (2) each convolution block extracts local and global temporal information with different feature map sizes; and (3) the extracted appearance and motion dynamic features are integrated through an injection from one to the other or with bi-direction way. 
More details about STFN are described in Section  \ref{STFN}.

\subsection{Spatio-temporal fusion networks} \label{STFN}

We consider the output feature maps of CNNs for $N$ segments from a video $V$. Each feature map $\{F_1, F_2, \cdots, F_N\}$ is a vector of size $F\in \mathbb{R}^d$, where $d$ is the output feature map dimension. The feature maps can be retrieved from different networks trained with different modalities such as appearance and motion. $F^a, F^m$, where $F^\mathrm{x} \in \mathbb{R}^{N\times d}$, are the feature maps from appearance and motion networks, respectively. STFN is applied to the sequence of feature maps, $F^a$ and $F^m$, to extract temporal dynamics of each feature map and fuse them as follows:
\begin{equation}
\begin{split}
\text{STFN}(F^a, F^m) &= \mathcal{H}(\mathcal{F}(\mathcal{G}(\mathcal{F}(F^a;\mathrm{W_a}),\mathcal{F}(F^m;\mathrm{W_m}));\mathrm{W_{fa}})) + \\ &\mathcal{H}(\mathcal{F}(\mathcal{G}(\mathcal{F}(F^a;\mathrm{W_a}),\mathcal{F}(F^m;\mathrm{W_m}));\mathrm{W_{fm}}))
\end{split}
\end{equation}

$\mathcal{F}(F^{\mathrm{x}};\mathrm{W_x})$, where $\mathrm{x} \in \{a,m,fa,fm\}$ meaning appearance, motion, fused appearance, fused motion sequences, is a ConvNet function with parameters $\mathrm{W_{\mathrm{x}}}$ which produces sequences of same input sizes for the given sequences. More details about the ConvNet are given in Section \ref{res-inc}. The fusion aggregation function $\mathcal{G}$ combines the output sequences of appearance and motion dynamic information. $\mathcal{G}$ and the follow-up ConvNets, $\mathcal{F}(F^{\mathrm{x}};\mathrm{W_{fa}})$, can be omitted depending on the design choice of STFN. More details are provided in the next subsection. From the learned sequences, the prediction function $\mathcal{H}$ predicts the probability of each activity class. Softmax function, which is widely used for multi-class classification, is chosen for $\mathcal{H}$. 
%The more detail information is provided in the next subsection.

The overall network is learned in an end-to-end scheme like TSN \cite{SOTA3}. The sequences of feature maps are $X={F^a, F^m}$ and the outputs of the $\mathcal{F}$ function are denoted by $y$. Also, let $\mathcal{L}$ be the loss function. The gradient of the loss function with respect to $X$, $\frac{d\mathcal{L}}{dX}$, during the training process is defined as:
\begin{equation}
\begin{split}
\frac{d \mathcal{L}}{d F^{\mathrm{x}}_k} = \mathcal{F}(F^{\mathrm{x}}_{k'};\mathrm{W_{\mathrm{x}}}) \frac{d \mathcal{L}}{dX}
\end{split}
\end{equation}

where $k \in N$ and $k'= \{1,2, \cdots, k-1, k+1, \cdots, N\}$. In the end-to-end training, the parameters for the $N$ segments are learned using stochastic gradient descent (SGD). The parameters are learned from the entire video with segmented temporal inputs.

\subsection{STFN components}

In this subsection, we describe the ConvNets, $\mathcal{F}$, and the fusion aggregation function, $\mathcal{G}$, in detail. We also discuss different STFN architectures to find the most suitable model.

\subsubsection{Residual Inception Block} \label{res-inc}

\begin{figure}[tb]
	\begin{center}
		\includegraphics[width=0.8\linewidth]{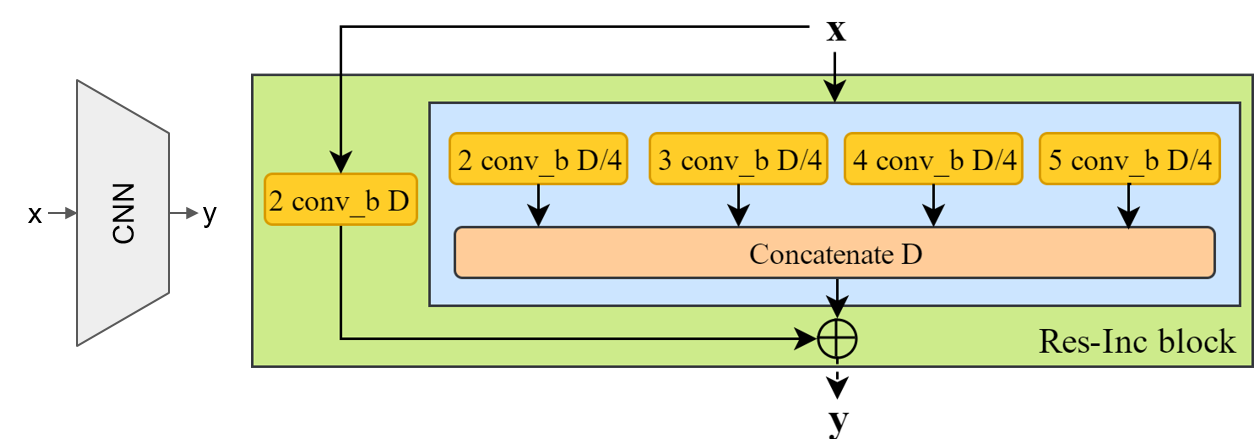}
	\end{center}
	\caption{
		A Residual Inception block. The res-inc block in the right figure shows the components of the CNN in the left figure. The number in each module inside of the Res-Inc block depicts convolution kernel size. conv{\_}b consists of the 1D convolution, batch normalization, and relu activation layers. $D$ represents the input vector dimension, $d$.}
	\label{fig:res_inception block}
\end{figure}

A sequence of frame representations, $F^a, F^m$, inherently contains temporal dynamics between features. The consecutive features are convoluted over time with different kernel sizes to extract local temporal information. This operation is conceptually similar to an n-gram of a sentence that contains local semantic information among n words. The convoluted features are then concatenated to formulate a hierarchical feature from each input. 
Motivated by an inception module \cite{ar_ref13,ar_ref14} that convolves an input signal with different filters, we design an inception block with different kernel sizes as shown in Fig.  \ref{fig:res_inception block}. 
The input signal $F^\mathrm{x}$ is convoluted across time using 1D convolution with four different sizes of kernels, 2,3,4,5, whose filter size is a quarter of the input dimension, $d$. The 1D convolution retains the same temporal length as the input. We did preliminary experiments to find out the best combination of the kernel sizes and 2,3,4,5 shows the best performance. We designed the filter size of each convolution to be a quarter of input dimension, making the concatenated feature have the same dimension as the input with same weight.
We also used convolution layers with kernel size of 1 \cite{ar_ref13,ar_ref14} before the conv{\_}b block to reduce the input dimension. However, they decrease the performance since it perturbs the input signal that contains temporal dynamics, so we decided not to include them.

%The filter size of the Res-Inc block for input and output signal used in our experiments is 2048. Note that an 1$\times$1 convolutions are used in  \cite{ar_ref13, ar_ref14} for the purpose of reducing or expanding the filter dimensions and making a network deep. However, a kernel size of 1 for the 1D convolution is not capable of incorporating temporal information from consecutive features. In this reason the size of 1 convolution filters are excluded in our design choice. 
The concatenated multi features and the input signal $F$ are added for residual learning \cite{ar_ref8}. 
%1$\times$1 convolution filter is optionally used in the shortcut connection \cite{ar_ref8} for the purpose of getting same number of channels but we choose a convolution filter size of 2 to capture local temporal information of the input signal. 
We chose a convolution kernel size of 2 for the skip connection to capture the smallest local temporal information.
Formally, the Residual Inception (Res-Inc) block in this paper is defined as:
\begin{equation}
\label{equ2}
\begin{split}
\mathrm{y} &= \mathcal{C}( \mathcal{R}\left(F^\mathrm{x}, \{W_i\}\right) )  + \mathcal{R}(F^\mathrm{x}, \{W_j\})	
\end{split}
\end{equation}
where $\mathcal{R}$ is the convolution function with weights $W_i, i \in \{2,3,4,5\}$ for the residual connection or $W_j, j \in \{2\}$ for the skip connection, and the function $\mathcal{C}(\cdot)$ represents a concatenation operation. In Fig. \ref{fig:res_inception block}, $\mathrm{x}$ is identical to $F^\mathrm{x}$ in Equation  \ref{equ2}.
The convolution block, conv{\_}b, is composed of Batch normalization \cite{ar_ref10} and ReLU \cite{ar_ref13}, while the convolution block in skip connection lacks the ReLU activation layer. The output signal is further activated with ReLU before it is aggregated with the other signal. 
%A more discriminative spatio-temporal features can be learned with a series of the Res-Inc blocks.
The output sequence of the Res-Inc block contains more discriminative temporal dynamic information than the input sequence. Since the Res-Inc block outputs signals of same dimension of input signals, a series of Res-Inc block can be easily setup.

\subsubsection{Spatio-temporal Fusion} \label{fusion_methods}

Despite the successful performance with the two-stream approach, a clear drawback is that a spatio-temporal information is not achievable with separate training of the appearance and motion data. The appearance and motion information are complementary to each other in order to discern an action of similar motion or appearance patterns e.g. brushing teeth and hammering. 
In order to overcome this deficiency, a number of researches have been looking into fusing two-stream networks \cite{st-arts4,st-arts5,ar_ref5} directly and learning spatio-temporal features \cite{ar_ref19,ar_ref21} . Although, their results show improved performance, their spatio-temporal features are limited to local snippets of an entire video sequence. In contrast, STFN takes advantage of extracted temporal dynamic features that capture long term temporal information over entire video to fuse them. 

We investigate three different fusion operations $\mathcal{G}$ with the output sequences of two Res-Inc blocks $\{P_1^\mathrm{x}, P_2^\mathrm{x}, \cdots, P_N^\mathrm{x}\}$, where $P_n^\mathrm{x} \in \mathbb{R}^d$ , and $\mathrm{x} \in \{a,m\}$ represent either appearance or motion features. 

\textbf{Element-wise Average}
\begin{equation}
\label{avg}
\begin{split}
P'_n = \frac{(P^a_n + P^m_n)}{2}
\end{split}
\end{equation}
where $P'$ is the aggregated sequence and $n \in \{1,2, \cdots, N\}$. This operation leverages all information and uses the mean activation for the fused signal. This operation may get affected by noisy input signals but since we deal with highly informative features, it is a good choice for our architecture.

\textbf{Element-wise Multiplication}
\begin{equation}
\label{mul}
\begin{split}
P'_n = P^a_n \times P^m_n
\end{split}
\end{equation}
The intuition behind this operation is to amplify a signal when both signals are strong, i.e. similar to attention mechanism. However, the noisy strong signal may affect heavily the fused signal leading to performance decrease.

\textbf{Element-wise Maximum}
\begin{equation}
\label{mul}
\begin{split}
P'_n = max(P^a_n ,P^m_n)
\end{split}
\end{equation}
The idea of max pooling is to seek the most discriminative signal among inputs. It selects either appearance or motion cue for each element of input signals. This operation may confuse the following Res-Inc block since the aggregated vectors are mixed with the appearance and motion signals.

We compare the performance of each operation in the ablation studies.

\begin{figure*}[!tbp]
	\centering
	\begin{subfigure}[b]{.31\textwidth}
		\centering
		\includegraphics[height=4.5cm, width=0.9\linewidth]{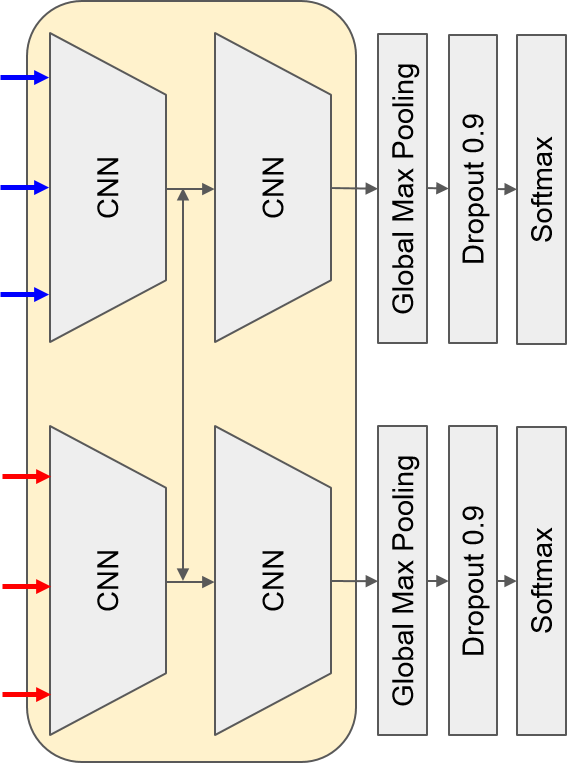}
		\caption{}
		\label{fig:network1}
	\end{subfigure}
	\begin{subfigure}[b]{.31\textwidth}
		\centering
		\includegraphics[height=4.5cm, width=0.9\linewidth]{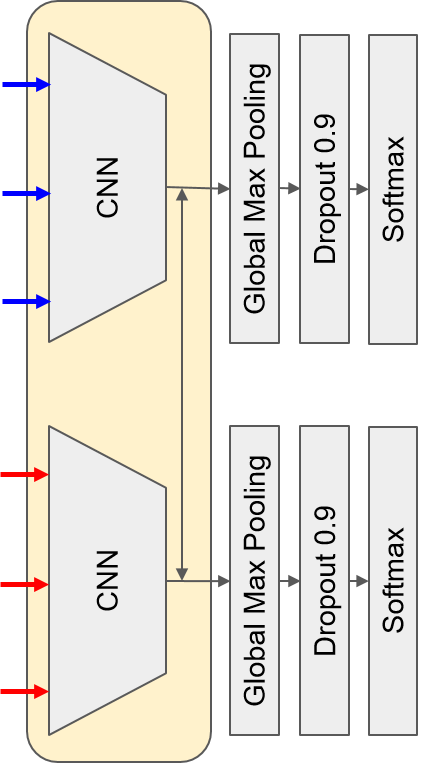}
		\caption{}
		\label{fig:network2}
	\end{subfigure}
	\begin{subfigure}[b]{.31\textwidth}
		\centering
		\includegraphics[height=4.5cm, width=0.8\linewidth]{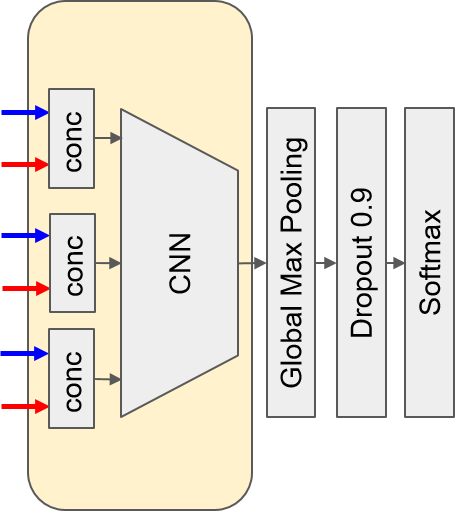}
		\caption{}
		\label{fig:network3}
	\end{subfigure}	
	\caption{
		Different designs of spatio-temporal fusion architecture. (a) shows our proposed architecture; (b) lacks the follow-up Res-Inc blocks after fusion; and (c) concatenation of the appearance and motion sequences in feature level before extracting temporal dynamics. The blue and red arrows represent the appearance and motion sequence inputs, respectively.
	}
	\label{fig:fusion_architecture}
\end{figure*}

\subsubsection{Architecture variations of STFN} \label{design_STFN}

\begin{figure}[tb]
	\begin{subfigure}[b]{.5\textwidth}
		\centering
		\includegraphics[width=0.7\linewidth]{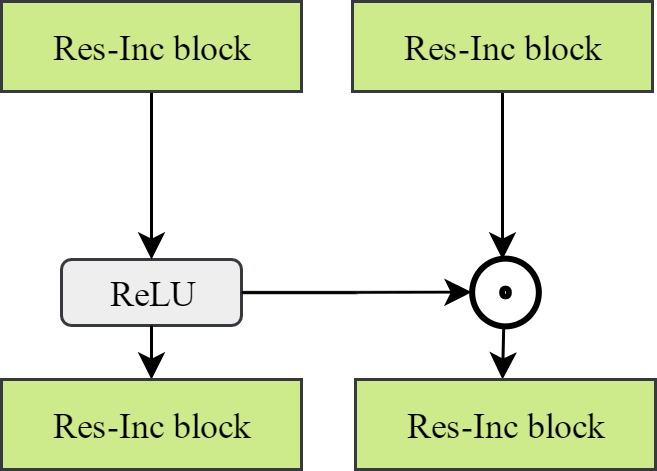}
		\caption{}
		\label{fig:asymetric}
	\end{subfigure}
	\begin{subfigure}[b]{.5\textwidth}
		\centering
		\includegraphics[width=0.7\linewidth]{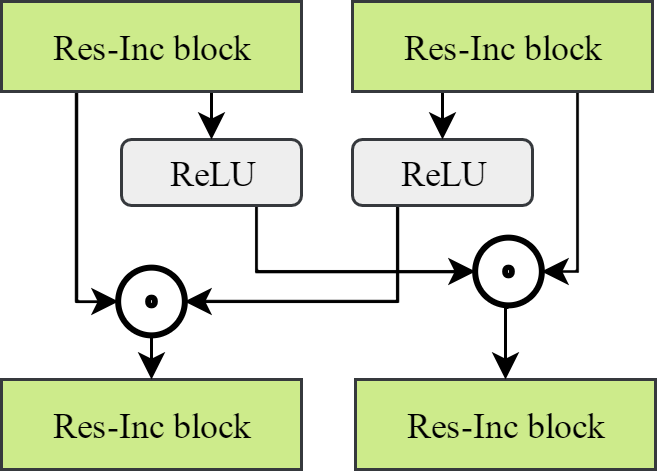}
		\caption{}
		\label{fig:bi-dir}
	\end{subfigure}		
	\caption{
		Two types of fusion methods: asymmetric and symmetric fusion. (a) shows asymmetric fusion method and two fusions are possible with this method: appearance to motion features and motion to appearance features. (b) shows symmetric fusion where each fused signal is further used in following layers. Two signals are merged with the previous described fusion operations. Note that this figure only illustrates the fusion connections between two Res-Inc blocks and the rest layers are omitted.
	}
	\label{fig:fusion_dir}
\end{figure}

We propose different design architectures of STFN and investigate them in detail. Fig. \ref{fig:network2} is a variation of Fig. \ref{fig:network1} where we want to learn how the additional Res-Inc blocks affect to the results. The Res-Inc blocks after fusion extract temporal dynamics of spatio-temporal features leading to better performance. In Fig. \ref{fig:network3}, fusion is executed in feature-level by simply concatenating appearance and motion signals. This fused signal is fed to the Res-Inc block to extract temporal dynamic information.

\subsubsection{Fusion direction} \label{fusion_direction}

As shown in Fig. \ref{fig:fusion_dir}, aggregating two signals can be three possible ways: appearance to motion, motion to appearance, and bi-directional fusion. The fused signals are fed to the next Res-Inc blocks and affect to the residual and skip connection along the forward an backward propagations when training. Considering the three fusion operations, only multiplication operation results in byproduct signal from partial derivatives of the fused signals when signals are back-propagated. This means the fusion with multiplication operation makes the input signal change rapidly than other operations.  Thus, it is not easy to learn proper spatio-temporal features especially when there is significant gap between the discriminative abilities of appearance and motion features.

%------------------------------------------------------------------------

\section{Experiments} \label{experiments}

In this section, we first discuss the datasets and implementation details. Then we evaluate each design choice for STFN. Finally, we compare our best performance with the state-of-the-art methods.

%The experimental results are provided with four sections. Sect. \ref{exp:fusion} presents ablation experiments on how to connect the two streams. Sect. \ref{exp:injection} shows the results using different integration methods and TSP. Sect. \ref{exp:network} describes the impact of employing deeper network. Lastly, Sect. \ref{exp:best} compares the state-of-the-art methods.

\subsection{Datasets}

We tested our method on two large action datasets, HMDB51 \cite{ar_ref3}, UCF101 \cite{ar_ref4}. The HMDB51 datset consists of 51 action classes with 6,766 videos and more than 100 videos in each class. All videos are acquired from movies or youtube and contain various human activities and interactions with human or object. Each action class has 70 videos for training and 30 videos for testing. 
The UCF101 dataset consists of 101 action categories with 13,320 videos and at least 100 videos are involved in each classes. UCF101 provides large diverse videos with a fixed resolution of $320\times240$ with 5 different types of actions. All videos are gathered from youtube.
Both datasets provide evaluation scheme for three training and testing splits and we follow the original evaluation method.

\subsection{Implementation details}

\textbf{Two-stream ConvNets}: 
ResNet-101 \cite{ar_ref8} and Inception-V3 \cite{ar_ref12} are employed for the base networks to train appearance and motion networks. Both networks are initialized with the pre-trained
weights trained on the  ImageNet \cite{dataset1} dataset. To fine-tune the networks, we replace the classification layer with $C$ softmax layer, where $C$ is the number of action classes. 
The appearance network takes RGB images, while the motion network a stack of 10 dense optical flow frames.
%We train the two streams separately as described in  \cite{ar_ref1}. 
%For the training of the spatial network, we did not use dropout layers. 
The input RGB or optical flow images are resized to make the smaller side as 256. We augment the input image by cropping, resizing, and mirroring in horizontal direction. The width and height of the cropped image are randomly sampled from $\{256,224,192,168\}$, and the input images are cropped from the four corners and the center of the original images. The cropped images are then resized to 224$\times$224 for the network input. This augmentation considers both scale and aspect ratio.
%For the training of the temporal network, we use dropout ratios of 0.8 after the final classification layer. 
We pre-compute the optical flows using the TVL1 method \cite{dataset2} before training to improve the training speed. The optical flow input is stacked with 10 frames making a $224\times224\times20$ sub-volume for $x$ and $y$ directions. Same data augmentation techniques are employed for the optical flow sub-volume. 
We use mini-batch stochastic gradient descent (SGD) to learn models with a batch size of 32 and momentum of 0.9. The learning rate is set to $10^{-3}$ initially and decreased by a factor of 10 when
the validation error saturates, for both networks.

\textbf{STFN}: 
In order to train STFN, we retain only convolutional layers and global pooling layer of each network, similar to  \cite{SOTA2}. The feature maps for STFN are extracted from the output of the global pooling layer. The output dimension is 2048 for both ResNet-101 and Inception-V3. We apply two step training process. We first fix the weights of trained appearance and motion networks and train only STFN. Then we train the entire networks with same methods described in Two-stream ConvNets training. For the first training, we initialize the learning rate with $10^{-4}$ and decrease it until $10^{-7}$ by a factor of 10 when the validation error saturates. RMSProp \cite{text_ref7} optimizer is used for the STFN training. The second training is executed with same setting of the two-stream ConvNets training without fixing all weights. For training and testing, we divide the videos into $N=5$ segments with same lenghts. Note that we use $N=5$ for all evaluation except for the experiment in Section \ref{exp:segments}. 
A random frame is selected from each $N$ segment and optical flow stacks centered on the selected frames are associated for two input sequences. We apply same augmentations for selected frames
and optical flow stacks in an input sequence. When testing, 5 frames are uniformly sampled from each segment making 5 sequences and the final prediction scores are averaged over each output. The experiments are performed with 5 segments, average fusion operation, and bi-directional fusion as defualt except for each ablation study.

\subsection{Evaluation of different designs}
\label{exp:design}

\renewcommand{\arraystretch}{0.9}
\begin{table}[tb]
	\begin{center}
		\begin{tabu} spread \linewidth {|X[c,m]|X[c,m]|X[c,m]|}
			\hline
			Design  & HMDB51 & UCF101 \\
			\hline
			\hline
			Fig. \ref{fig:network1} & \textbf{70.4} & \textbf{93.5}\\
			\hline
			Fig. \ref{fig:network2} & 69.6 & 93.2  \\
			\hline			
			Fig. \ref{fig:network3} & 69.2 & 92.0  \\
			\hline						
		\end{tabu}
	\end{center}
	\caption{
		Prediction accuracy($\%$) on the first split of HMDB51 and UCF101 using different architectures of STFN as shown in Fig. \ref{fig:fusion_architecture}.
	}	
	\label{table:design}
\end{table}

As we discussed in Section \ref{design_STFN}, the performances of three proposed STFN architectures are presented in Table \ref{table:design}. Comparing Fig. \ref{fig:network1} and Fig. \ref{fig:network2} networks, we verify that the Res-Inc blocks make important role extracting temporal dynamics. We conjecture that the consecutive Res-Inc blocks extract temporal dynamics of fused features and they contain better video-wide discriminative features. Another architecture design, Fig. \ref{fig:network3}, is introduced to see how the feature level fusion affects to the performance as opposed to the baseline two-stream networks. We observe the significant performance drop in both datasets and it proves the importance of the fusion scheme. 
%The experiments are performed with 5 segments, average fusion operation, and bi-directional fusion. 
Since the architecture of Fig. \ref{fig:network1} shows the best performance, we choose it as our default STFN network. 

The result with a single Res-Inc module (\ref{fig:network2}) outperforms the baseline late fusion results shown in Table \ref{table:baseline} by 8.1$\%$ on HMDB51 and 0.2$\%$ on UCF101. This shows the effectiveness of the Res-Inc module.  With another Res-Inc module and feature fusion, 0.8$\%$ and 0.3$\%$ additional gains are obtained on HMDB51 and UCF101, respectively.
Note that from a preliminary experiment by increasing the number of consecutive Res-Inc blocks from two to four, we observe performance drops: 3.8$\%$, 6.5$\%$ on HMDB51, 4.1$\%$, 6.9$\%$ on UCF101. The signals undergo the Res-Inc block contain temporally convoluted information with different kernel sizes (from residual connection). More Res-Inc blocks extract higher level temporal information, but we conjecture that signals experienced more than two levels confuse the original temporal orders, introducing noise.

\subsection{Evaluation of fusion operations}
\label{exp:fusion_operation}

\renewcommand{\arraystretch}{0.9}
\begin{table}[tb]
	\begin{center}
		\begin{tabu} spread \linewidth {|X[c,m]|X[c,m]|X[c,m]|}
			\hline
			Fusion operation  & HMDB51 & UCF101 \\
			\hline
			\hline
			Average &  \textbf{70.4} & \textbf{93.5}\\
			\hline
			Maximum & 69.5 & 92.9  \\
			\hline			
			Multiplication & 68.3 & 92.6  \\
			\hline						
		\end{tabu}
	\end{center}
	\caption{
		Prediction accuracy($\%$) on the first split of HMDB51 and UCF101 using different fusion operations.
	}	
	\label{table:fusion_operation}
\end{table}

This section presents the performances based on different fusion operations: Element-wise average, maximum, and multiplication. As shown in Table \ref{table:fusion_operation}, the average operation outperforms other methods. It is interesting to see the performance gap between the average and the multiplication operations, 0.9$\%$ and 2.1$\%$ for HMDB51 and UCF101 respectively. We speculate the reason is due to the performance discrepancy of two networks as shown in Table \ref{table:baseline}. With multiplication, the inferior feature (appearance cue on HMDB51) could harm the fused signal. Also, it is better to take into account all data by averaging than picking the strongest signals since STFN deals with highly pre-processed signals. From the results, we take the average operation as our default choice. Note that we tried weighted average based on the normalized performances of baseline networks and automatic scaling by appliying 1x1 2D conv to each signal before fusing. However simple average results in the best performance.

\subsection{Evaluation of fusion directions}
\label{exp:fusion_direction}

\renewcommand{\arraystretch}{0.9}
\begin{table}[tb]
	\begin{center}
		\begin{tabu} spread \linewidth {|X[c,m]|X[c,m]|X[c,m]|}
			\hline
			Fusion direction  & HMDB51 & UCF101 \\
			\hline
			\hline
			A$\leftrightarrow$M & \textbf{70.4} & \textbf{93.5}  \\
			\hline						
			A$\leftarrow$M &  70.3 & 93.4\\
			\hline
			A$\rightarrow$M & 70.1 & 93.2  \\
			\hline			
		\end{tabu}
	\end{center}
	\caption{
		Prediction accuracy($\%$) on the first split of HMDB51 and UCF101 using different fusion directions. A and M represent the appearance and motion features, respectively. The bottom two methods are asymmetric fusion methods whereas the top one is bi-direction fusion method.
	}	
	\label{table:fusion_direction}
\end{table}

%We discuss six different fusion methods in Sect. \ref{fusion_methods}. There are many other ways to connect two or more networks but we investigate all possible fusion methods in terms of variation of applying the non-linearity activation to the signals. 
In Table \ref{table:fusion_direction}, we compare the performance variation with different fusion directions.  Note that A$\leftarrow$M is a simply reflected network of A$\rightarrow$M and we use 5 segments for all experiments. 
For the asymmetric fusion methods, A$\leftarrow$M connection outperforms the other way consistently on both datsets. This effect is due to the fact that the motion stream overfits quickly with the A$\rightarrow$M fusion and no further spatio-temporal learning occurs. This comes from the base performance different between appearance and motion features so that fusion injection to the higher discriminative feature leads to worse performance. 
The bi-direction fusion outperforms A$\leftarrow$M with small margin, 0.1$\%$ on HMDB51 and 0.2$\%$ on UCF101. This makes sense since two spatio-temporal features are learned simultaneously in two streams, whereas asymmetric fusion learns spatio-temporal in the injected stream and the learned weights are propagated to the other stream only when back propagating from the fused connection. However, we argue that our proposed STFN is robust to the fusion connection based on the small performance differences on both datasets. We choose the bi-direction fusion as our base fusion method.

\renewcommand{\arraystretch}{0.9}
\begin{table}[tb]
	\begin{center}
		\begin{tabu} spread \linewidth {|X[1c,m]|X[c,m]|X[c,m]|}
			\hline
			Number of segments  & HMDB51 & UCF101 \\
			\hline
			\hline
			3 & 70.3 & 93.2  \\
			\hline						
			5 &  70.4 & 93.5\\
			\hline
			7 & \textbf{70.8} & \textbf{93.9 } \\
			\hline			
			9 & 70.5 & 93.6  \\
			\hline			
		\end{tabu}
	\end{center}
	\caption{
		Prediction accuracy($\%$) on the first split of HMDB51 and UCF101 using different numbers of segments in videos.
	}	
	\label{table:segments}
\end{table}

\subsection{Evaluation of a number of segments}
\label{exp:segments}

We evaluate the number of segments according to the default fusion method and architecture. One may assume that more segments result in better performance. However, as we discussed, more redundant temporal dynamics are introduced when increasing the number of segments.
The performances based on different number of segments are shown in Table \ref{table:segments}. It turns out that 7 segments performs best and 0.4$\%$ performance increases are observed on both datasets. 
The STFN with 9 segments underperforms compared with the one with with 7 segments. We verify our hypothesis with this experiments that sparse sampling is necessary to avoid redundant temporal dynamics over entire videos. For the best network, we determine the number of segments as 7.

\subsection{Base performance of two-stream network}
\label{exp:network}

\renewcommand{\arraystretch}{0.9}
\begin{table}[!tb]
	\begin{center}
		\begin{tabu} {|X[c,m]|X[1.1,c,m]||X[c,m]|X[c,m]|X[c,m]|}
			\hline
			Dataset & Network & Appear. & Motion & Late Fusion \\			
			\hline
			\hline
			\multirow{2}{*}{HMDB51} & ResNet-101 & 48.2 & 58.1  & 61.1 \\
			\cline{2-5}
			& Inception-V3 & 51.2 & 59.2 & 62.7\\			
			\hline
			\multirow{2}{*}{UCF101} & ResNet-101 &83.5 & 86.0 & 91.8\\
			\cline{2-5}
			& Inception-V3 & 84.8 & 88.1 & 92.3\\
			\hline
		\end{tabu}
		%\end{tabular}		
	\end{center}
	\caption{
		Performance comparison($\%$) of two-stream networks with ResNet-101 and Inception-V3 on HMDB51 and UCF101 (split1). Inception-V3 shows consistently better prediction accuracies over ResNet-101 on both appearance and motion networks.
	}
	\label{table:baseline}
\end{table}

We compare the different ConvNet architectures for STFN. ResNet-101 \cite{ar_ref8} and Inception-V3 \cite{ar_ref12} networks are employed to train the two-stream networks. As shown in Table \ref{table:baseline}, the performance with Inception-V3 is better than ResNet-101 on both datasets. The performance gaps of the appearance and motion networks are 3.0$\%$/1.1$\%$ on HMDB51 and 1.3$\%$/2.1$\%$ on UCF101, respectively.

\subsection{Comparison with the State-of-the-art}
\label{exp:best}

We compare STFN with the current state-ot-the-art methods in Table \ref{table:state_art}. We report the mean accuracy over three splits of the HMDB51 and UCF101.
The first section of Table \ref{table:state_art} consists of the hand-crafted features with different encoding methods. The second and third sections describe approaches using ConvNets but the methods in third section utilize additional modalities for the final prediction.
STFN with the Inception-V3 achieves the best results: 72.1$\%$ on HMDB51 and 95.4$\%$ on UCF101. There is 0.9$\%$/1.1$\%$ performance increase from STFN with ResNet-101 architecture. STFN with both networks shows the state-of-the-art performance.
Comparing with baseline late fusion performance of two-stream networks, performance increases are observed as follows: 9.4$\%$, 10.1$\%$ on HMDB51 and 3.1$\%$, 2.5$\%$ on UCF101 with Inception-V3 and ResNet-101, respectively. 

Our best results outperform TSN \cite{SOTA3} by 1.0$\%$ on HMDB51 and 0.5$\%$ on UCF101 with same number of segments, 7. While TSN predicts scores with consensus operations and averages each score, STFN extracts temproal dynamic information and aggregates signals in feature level leading to better results. 
The results prove our method produces effective spatio-temporal features.
DOVF \cite{SOTA6} and TLE \cite{SOTA2} show better results than STFN with ResNet-101 but are outperformed by STFN with Inception-V3. 
TLE \cite{SOTA2} only outperforms our method with small margin, 0.2$\%$, on UCF101 but the gap is reversed with additional hand-crafted feature score.

We combine our results with the hand-crafted MIFS\footnote{The prediction scores of MIFS are downloaded from \href{http://www.cs.cmu.edu/~lanzhzh/}{HERE}.} \cite{feat_ref5} features by averaging prediction scores. The performance gain on HMDB51, 3,0$\%$, is larger than on UCF101, 1.6$\%$. The combined performances, 75.1$\%$ on HMDB51 and 96.0$\%$ on UCF101, outperform all state-of-the-arts and even on par with  \cite{SOTA7,SOTA4} which employ more prediction scores from additional modalities. Note that we observe similar performance boost with iDT \cite{feat_ref3} but choose MIFS since the prediction scores are available in public.

\renewcommand{\arraystretch}{1.}
\begin{table}[!tb]                             
	\centering                                
	%\begin{tabular}{|c|c|c|c|}                
	\begin{tabu} to 1.0\linewidth {|X[1.1,c,m]||X[c,m]|X[c,m]|}  
		\hline 
		%\multicolumn{2}{|c|}{HMDB51} & \multicolumn{2}{c|}{UCF101}\\                  
		Methods & HMDB51 & UCF101 \\
		\hline \hline                              
		iDT+FV \cite{ar_ref16} & 57.2 &  85.9 \\          
		\hline                                    
		iDT+HSV \cite{bovw_ref8} & 61.1  & 87.9 \\            
		\Xhline{0.7pt}
		\Xhline{0.7pt}
		Two-stream \cite{ar_ref1} & 59.4 & 88.0 \\
		\hline
		%C3D+iDT \cite{ar_ref19} & $\textendash$ & 90.4\\
		%\hline
		%TDD+FV \cite{ar_ref7} & 63.2 & 90.3 \\             
		%\hline                                    
		Transformation \cite{ad_ref4} & 62.0 & 92.4 \\ 
		\hline
		KVM \cite{ar_ref11} & 63.3 &  93.1 \\              
		\hline                                    
		Two-Stream Fusion \cite{ar_ref5} & 65.4 (69.2 i) &  92.5 (93.5 i) \\
		\hline
		ST-ResNet \cite{st-arts5} & 66.4 (70.3 i) & 93.4 (94.6 i) \\ 
		\hline		
		ST-Multiplier \cite{st-arts4} & 68.9 (72.2 i) & 94.2 (94.9 i)\\ 
		\hline						
		ActionVLAD \cite{st-arts6} & 66.9 (69.8 i)& 92.7 (93.6 i) \\ 
		\hline		
		ST-Vector \cite{st-arts7} & 69.5 (73.1 i+H) & 93.6 (94.3 i+H)\\ 		
		\hline		
		DOVF \cite{SOTA6} & 71.7 (75.0 M) & 94.9 (95.3 M)\\ 
		\hline		
		ST-Pyramid \cite{SOTA1} & 68.9 & 94.6\\ 
		\hline
		I3D \cite{st-arts8} & 66.4 & 93.4 \\
		\hline		
		CO2FI \cite{SOTA5} & 69.0 (72.6 i) & 94.3 (95.2 i)\\ 		
		\hline		
		TLE \cite{SOTA2} & 71.1 & \textbf{95.6}\\ 		
		\hline				
		TSN \cite{SOTA3} & 71.0 & 94.9 \\ 
		\Xhline{0.7pt}
		\Xhline{0.7pt}
		Four-Stream \cite{SOTA7} & 72.5 (74.9 i) & 95.5 (96.0 i)  \\ 
		\hline
		OFF \cite{SOTA4} & 74.2 & 96.0  \\ 		
		\Xhline{0.7pt}
		\Xhline{0.7pt}
		STFN (ResNet-101) & 71.2 (73.3 M) & 94.3 (95.1 M) \\
		\hline                                    
		STFN (Inception-V3) & \textbf{72.1} (\textbf{75.1} M)  & \textbf{95.4} (\textbf{96.0} M) \\
		\hline                                    
	\end{tabu}                             
	\caption{
		Comparison with state-of-the-art methods on HMDB51 and UCF101. Mean accuracy over three splits. Numbers inside of parenthesis are classification accuracies with hand-crafted features. (i: iDT \cite{feat_ref3}, H: HMG \cite{feat_ref6}, M: MIFS \cite{feat_ref5})
	}
	\label{table:state_art}   
	%\vspace*{-0.5cm}	                    
\end{table} 

%------------------------------------------------------------------------
\section{Conclusion} \label{conclusion}

In this paper, we introduced the sptio-temporal fusion network (STFN), a network suitable for extracting temporal dynamics of features and learning spatio-temporal features by combining them.
The spatio-temporal features are learned  effectively with STFN via an end-to-end learning method. 
%With the architecture, spatio-temporal  learns how spatial and temporal features change over time and fuses them for better video level representation. 
In the ablation studies, we show the best fusion methods and architecture and investigate the intuition behind each method.
STFN enables appearance and motion dynamic features integrate inside of the networks in a highly abstract manner and overcomes the naive fusion strategy of late fusion.
STFN is applicable to any sequencial data with two different modalities and effectively fuses them into highly discriminative feature that captures dynamic information over the entire sequence.
The best result of STFN achieves the state-of-the-art performance, 75.1$\%$ on HMDB51 and 96.0$\%$ on UCF101.
As future work, we consider scalability of our work with larger dataset and applying more than two modalities.

%===========================================================
%\tiny
%\scriptsize
%\footnotesize
%\small
%\normalsize

\bibliographystyle{splncs04}
\bibliography{bib} % \scriptsize

\end{document}